\documentclass[]{article}
\usepackage[utf8]{inputenc}
\usepackage{float}
\usepackage{amsmath}
\usepackage{amsfonts}
\usepackage{amssymb}
\usepackage[margin=2.5cm]{geometry}
\usepackage{natbib}
\usepackage{graphicx}
\usepackage{hyperref}

\usepackage{booktabs}
\usepackage{authblk}
\title{Local Dynamic Modes of Cognitive Behavioral Therapy}
\author[1]{Victor Ardulov\footnote{corresponding author: \url{ardulov@usc.edu}}}
\author[2]{Torrey Creed}
\author[3]{Dave Atkins}
\author[1]{Shrikanth Narayanan}

\affil[1]{University of Southern California}
\affil[2]{University of Pennsylvania}
\affil[3]{University of Washington}

\date{April 2022}

\begin{document}

\maketitle

\abstract{In order to increase mental health equity among the most vulnerable and marginalized communities, it is important to increase access to high-quality therapists. One facet of address these needs, is to provide timely feedback to clinicians as they interact with their clients, in a way that is also contextualized to specific clients and interactions they have had. Dynamical systems provide a framework through which to analyze interactions. The present work applies these methods to the domain of automated psychotherapist evaluation for Cognitive Behavioral Therapy (CBT). Our methods extract local dynamic modes from short windows of conversation and learns to correlate the observed dynamics to CBT competence. The results demonstrate the value of this paradigm and outlines the way in which these methods can be used to study and improve therapeutic strategies.}
% \keywords{Behavioral Signal Processing, Cognitive Therapy, Dynamical Systems}

\maketitle

\section{Introduction}
As mental health awareness and diagnosis become more pervasive so does the need for access to high quality counseling and psychotherapy. As of \citeyear{abuse2019mental} about 1 in 5 adults in the United States (19\%) reported to be diagnosed with a mental, behavioral, or emotional disorder and half as many, 9.5\% of adults in the US, report receiving counseling\citep{abuse2019mental, terlizzi2020mental}. In order to sustainably and equitably provide access to care it is imperative that therapists and clinicians are equipped with technologies and tools that best support their skills.  

% Recent efforts to increase equtible access to effective mental health care  have resulted in policy shifts toward evidence-based treatments, or the treatments that have shown consistently strong outcomes in clinical research. Initiatives such as those described in Creed et al. 2016 demonstrate that these evidence-based treatments, such as Cognitive Behavioral Thearpy (CBT), can also be effectively delivered outside of the controlled research environment in contexts like community mental health care settings. In order to support high-quality care and support positive clinical outcomes, researchers developed the Cognitive Therapy Rating Scale (CTRS; Young & Beck, 1980) to assess a therapist’s skill in delivering the core components of CBT. The use of the CTRS during training and evaluation provides concise and targeted feedback to therapists about how they might improve their CBT delivery, which in turn has been shown to improve clinician competence in the therapy they deliver to clients (Creed et al, 2016).”
Recent efforts to increase equitable access to effective mental health care have resulted in policy shifts towards treatments with consistent evidence of strong outcomes in clinical research. Initiatives such as those described in \citet{creed2016implementation} demonstrate how these treatments, for Cognitive Behavioral Therapy (CBT), can be successfully delivered outside of clinical research environments, including community mental health care settings. In order to support high-quality care and support positive clinical outcomes, researchers and clinicians developed the Cognitive Therapy Rating Scale (CTRS) to assess a therapists' competence to core components of CBT \citep{young1980cognitive}. During periods of evaluation and training, the use of CTRS provides therapists with concise and targeted feedback to suggest adjustments to improve their CBT delivery. This feedback in turn has been shown to improve consistency of clinician competence in the therapy they deliver to clients \citep{creed2016implementation}.

A critical limiting factor in scaling these initiatives is the need for expert evaluations to assess therapists via CTRS, as these evaluations are quite time-critical and resource intensive. By increasing access to timely and concise feedback, therapists would be able to strengthen their clinical care delivery. This in turn leads to a higher bandwidth of support for clients, by enabling more professionals to be available, and improve overall quality of care. Towards this end, previous works in automated psychotherapy assessment and feedback have demonstrated success in data-driven evaluation of therapist competence \citep{gibson2019multi, chen2021feature, flemotomos2018language, flemotomos2021automated}. These works utilize acoustic and language elements of speech towards predicting behaviors and outcomes of the clients and the therapists' utilization of therapeutic skills. 

The presented work outlines an alternative paradigm through which to study the interaction, namely by re-imagining the interaction as a dynamical system it is possible to construct a framework under which the flow of conversation can be studied as a control-affine system. This paradigm would enable the optimization of therapeutic strategies, as well as, have directly interpretable model parameters. The approach presented will construct and fit local dynamical systems models over short windows of interaction, extracting the dynamic modes via eigenvalue decomposition and demonstrate that these modes carry information that is pertinent to the competence of CBT tenants. Our results suggest that there is a natural interpretation of these modes as momentary indicators of strategies that are informative for assessing therapists' competence.

\subsection{Prior Work}
\subsubsection{Automated Therapy Evaluation}
\label{sec:automate_therapy_evaluataion}
The process of identifying patterns indicative of therapy quality and client outcome requires domain expertise, and is cognitively straining and time-consuming. To help overcome these limitations and support practitioners, behavioral signal processing (BSP) has been introduced as a context to utilize data towards automating many of the involved tasks \citep{narayanan2013behavioral}. 

An early demonstration of how data-driven methods could support automated psychotherapy evaluation, \citet{gibson2016deep} introduced the processing of dialogue transcripts with deep recurrent neural networks (RNNs) to predict session-level empathy ratings for therapists practicing Motivational Interviewing (MI). \citet{gibson2017attention} built on this work by introducing attention mechanisms on top of RNNs for automatically identifying MI skills on an utterance level. These results demonstrate how behavioral signal data, acoustic and linguistic time-series, could be evaluated to construct representation of nuanced dialogue acts. By predicting local actions taken by the therapist, these models gave way for evaluating strategies on a more generalizable level, for example allowing the evaluation of different question types on therapeutic outcomes. Most recently \citet{flemotomos2021amiagood} described a complete pipeline which is able to translate from raw audio to full transcripts, with utterance and session level predictions for MI behavioral skills and assessment ratings. These pipelines demonstrate the power of providing feedback in a timely manner, and open a natural opportunity for continued learning and refinement.

Specifically in the domain of CBT evaluation, \citet{flemotomos2018language} first compared different language models for evaluating the CTRS scores, while \citet{gibson2019multi} utilized multi-task learning to improve automatic CTRS assessment by utilizing experiences from other therapy styles such as MI. \citet{chen2021feature} evaluated the value of enriching language features with behavioral skill codes and dialogue acts. Recently, \citet{flemotomos2021automated} demonstrated the use of utilizing fine-tuned large language models to improve the discrimination of high- and low-competence CBT strategies.

Prior work has demonstrated how the analysis of speech signals and contents of clients' and therapists' language can be powerful indicators of what strategies therapists are using and the overall quality of the session with respect to clinically relevant metrics. In contrast to previous works, the presented study specifically evaluates the dynamics of therapeutic dialogue as observed from language. The methodology outlined constructs a more general framework for the study of interpersonal interactions and demonstrates how these models capture pertinent dynamical modes which can be translated into meaningful CBT strategies. 

In contrast to the aforementioned deep neural methods, our approach looks at more explicitly at \textit{control-affine dynamical system} models. This model applies an interpretation to the observed data as a system in which the output of the client is result of a previous state that is undergoing a self-regulating process (\textit{transition}) that is influenced by the observed data from the therapist (\textit{control}) at a given time point. This coupled pair of transition and control is referred to the dynamics of the system as they describe the evolution of the interaction instead of the \textit{static} content of the interaction. Rather than fitting a single model to all of the data available, each interaction is fit separately and then the parameters of the fit model can be evaluated directly via methods such as eigenvalue decomposition.

The methods, outlined in more detail in \S \ref{sec:windowed_dmd}, have been used to extract behavioral dynamics to study Child Forensic Interviews \citep{ardulov2018multimodal, durante2022causal} and infant-mother interaction modeling \citep{klein2021dynamic} demonstrating that interpersonal interactions could be holistically evaluated under with these considerations.
 
\begin{table}[ht]
    \centering
    \begin{tabular}{c|c}
        \toprule
        Score & Abbreviation \\
        \midrule
         agenda &  ag\\
         application of technique & ap \\
         collaboration & co \\
         feedback & fb \\
         guided discovery & gd \\
         homework & hw \\
         interpersonal & ip \\
         key cognition behavior & cb \\
         pacing and timing & pt \\
         strategy for change & sc \\
         understanding & un \\
         total CTRS & ctrs\\
         \bottomrule
    \end{tabular}
    \caption{Abbreviations of sub-scores}
    \label{tab:subscore_abbreviation}
\end{table}
\section{Methods}
CBT is focused on addressing mental health problems through the conscious exercise of cognitive change strategies \citep{beck2011cognitive, creed2016implementation}, outlining techniques that the therapist conversationally guides their client through. The therapist engages the client in a series of activities to help them build the necessary skills to shift their patterns of thinking and reacting to situations. The procedure highlights 11 elements that a therapist should be integrating throughout the session, such as setting an agenda, providing feedback, or utilizing homework, which are then summarily scored together, as described in Table \ref{tab:subscore_abbreviation}.

The present work is focused on the identification of types of dialogue flows that demonstrate competence in CBT practices. Our approach evaluates the likelihood that the extracted behavioral dynamics occur in a session where the therapist's skills are delivered effectively. 
Under the control-affine system paradigm the therapists' utterances are considered as an input signal to the system described by the present client, and the clients' utterances are viewed as the observable output from the system, similar to how a sensor might be used to estimate the state of a system. This perspective explicitly models the client as the central view of the dynamics. More concretely, the dynamics analyzed in this methodology are a combination of the observable evolution of the client's behaviors as reflected by their utterances in response to the therapists input. 

In \S \ref{sec:data}, the data and the steps taken to process the data into the format necessary to conduct the study. Next, \S \ref{sec:baselines} outlines the two classification tasks that are considered and the baselines against which the methods are considered, while \S \ref{sec:windowed_dmd} presents the models used for the tasks. Finally, \S \ref{sec:results} demonstrates the results, highlighting the effectiveness of the approach to simultaneously identify local interactions that are indicative of high session quality and discusses the conclusions which can be drawn.

\begin{figure}[ht]
    \centering
    \includegraphics[width=0.95\linewidth]{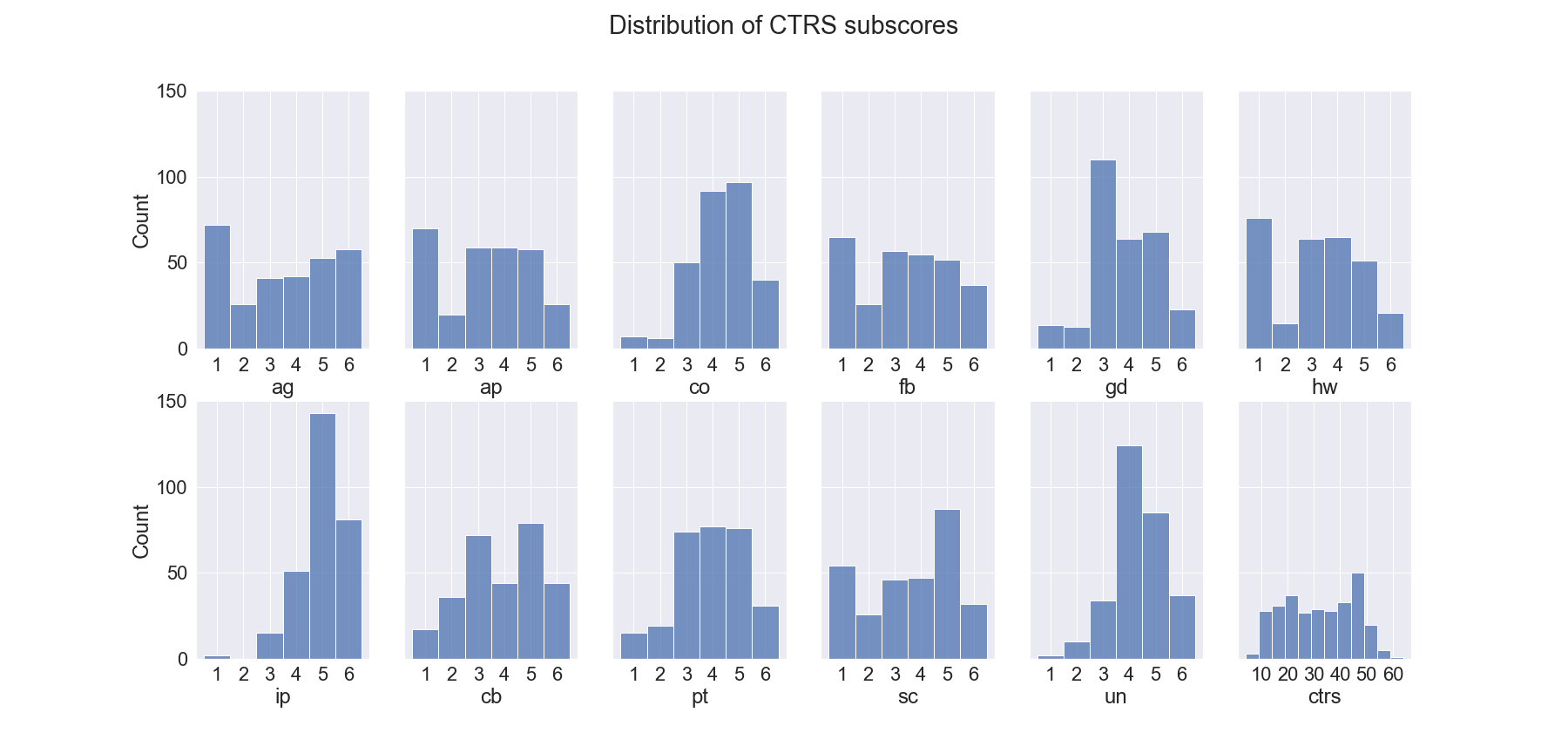}
    \caption{Distribution of the different CTRS subscores and the total CTRS scores in the data}
    \label{fig:score_distributions}
\end{figure}

\subsection{Data}
\label{sec:data}
The present study is conducted by analyzing the transcripts of 292 recorded CBT sessions with 175 unique clients. The transcripts are annotated by professionally trained coders for each CTRS sub-score, which in turn are aggregated to compute a total score. The individual sub-scores, named explicitly in Table \ref{tab:subscore_abbreviation}, are each scored on a scale between 0 and 6. Having a skill rated as a 4 or higher is considered demonstrating high-competence, and subsequently a total CTRS score of 40 is the threshold necessary for a session to be considered as high-competence \citep{vallis1986cognitive}. These thresholds outlined by the CTRS coding manual were in turn converted into binary labels where 0 and 1 represented low- or high-competence respectively. Figure \ref{fig:score_distributions} can be evaluated for the distribution of scores for each of the sub-scores and the summary CTRS score.
 
\begin{figure}[ht!]
    \centering
    \includegraphics[width=0.65\linewidth]{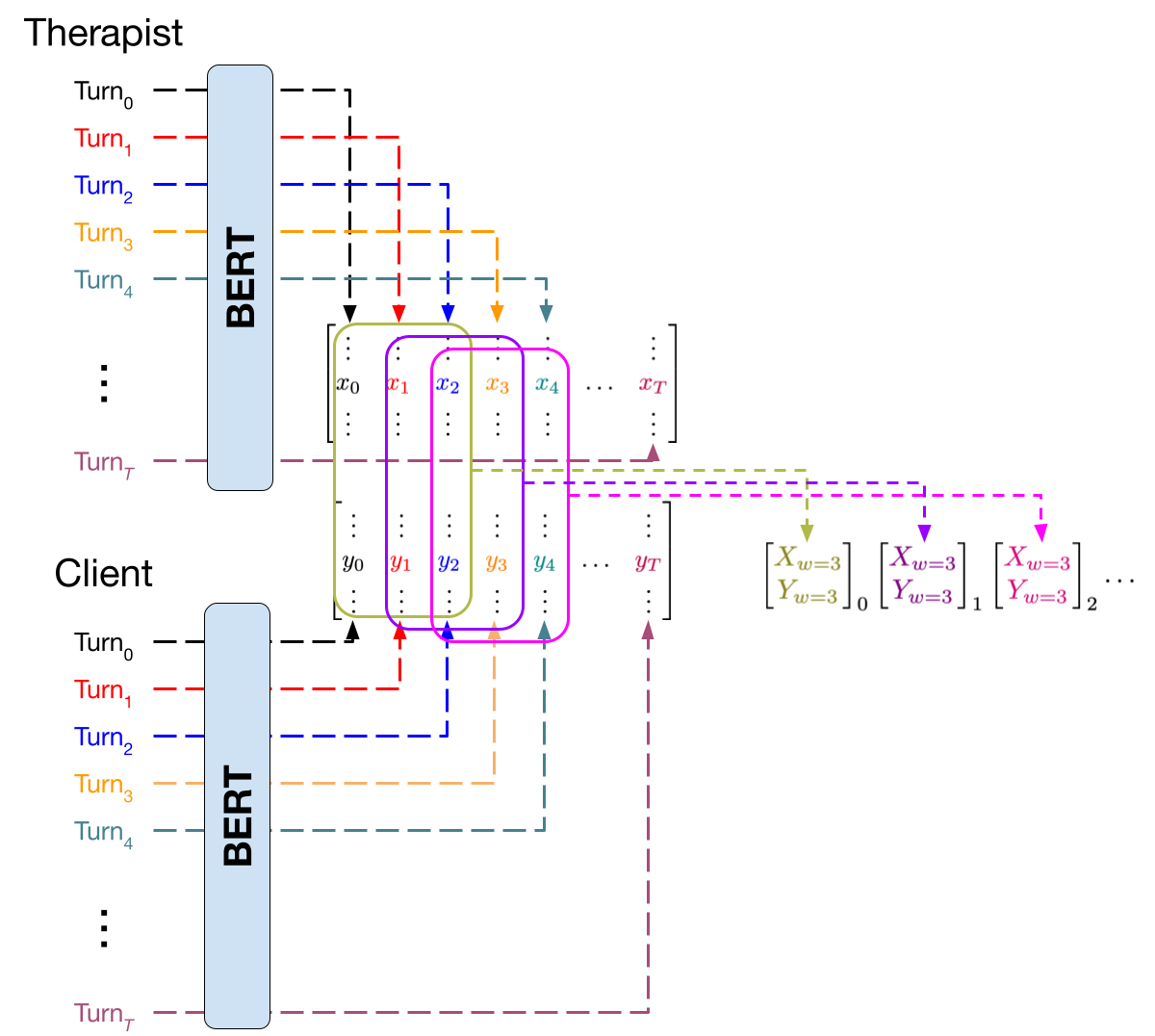}
    \caption{Each talk turn is converted into their respective vector representation via the DistilBERT transformer model, and the used to construct $X$ and $Y$ matrices, which are in turn window conditioned on the window size $w$.}
    \label{fig:window_extraction_pipeline}
\end{figure}

For each of the transcripts, the talk turns were converted into embeddings using a transformer based language model such as those described in \citet{devlin2018bert}. In this way, these embeddings can be considered the inputs and observables of the dynamical system model where each talk turn is considered a time-step.\footnote{Embeddings were extracted using the DistilBERT model found at \url{https://huggingface.co/distilbert-base-uncased}} An illustration of the process taken to convert talk turns into aligned embedding can be found in Figure \ref{fig:window_extraction_pipeline}, which also demonstrates how the data were grouped into windows which would be the inputs into the models described in \S \ref{sec:windowed_dmd}.

\subsection{Tasks and Baselines}
\label{sec:baselines}
Using the labels outlined in \S \ref{sec:data}, our approach is interested in associating windows of dialogues with whether or not they belong to a CBT competent session. To accomplish this two classification tasks are considered. The first, referred to as the \textit{local} task, looks at taking a window of talk turns and predicting the global binary label given to the session as a whole. Presumably, there will be similar windows of dialogue that occur during both high- and low-competence sessions since they are not typically completely absent of strategies that are also employed in high-competence settings, rather low-competence sessions typically under deliver these strategies. For this reason the models' output is considered to be a probability that the window came from a high-competence session, consequentially providing a discriminator that can be aggregated over the session. Since the underlying hypothesis is that a high-competence session will have more windows of high-competence behavior than the low-competence sessions, the second \textit{global} classification task, takes aggregate predictions over the entire session and predicts the final binary label using the combined score. 
For each of the \textit{local} and \textit{global} tasks, bootstrapped simulated baselines were constructed. Since the size of the window impacts the number of samples available to predict over, separate baselines were computed for each size in the \textit{local} scenario. The \textit{local} $3\sigma$ performance was used as a threshold to consider a model for the global task. Bootstrap F1 results can be found in Tables \ref{tab:local_three_sigmas} and \ref{tab:global_3_sigmas}, which display thresholds of significance which are used to evaluate the models against. 

\begin{table}[ht!]
\small
\centering
    \begin{minipage}{0.4\linewidth}
        \centering
        \begin{tabular}{lcc}
        \toprule
            Score &  Window Size &  F1 \\
        \midrule
            ag &            3 &    0.5115 \\
            ag &            5 &    0.5117 \\
            ag &            8 &    0.5118 \\
            \midrule
            ap &            3 &    0.5122 \\
            ap &            5 &    0.5124 \\
            ap &            8 &    0.5126 \\
            \midrule
            co &            3 &    0.5121 \\
            co &            5 &    0.5122 \\
            co &            8 &    0.5123 \\
            \midrule
            fb &            3 &    0.5082 \\
            fb &            5 &    0.5083 \\
            fb &            8 &    0.5084 \\
            \midrule
            gd &            3 &    0.5128 \\
            gd &            5 &    0.5129 \\
            gd &            8 &    0.5131 \\
            \midrule
            hw &            3 &    0.5101 \\
            hw &            5 &    0.5102 \\
            hw &            8 &    0.5104 \\
            \bottomrule
            \end{tabular}
    \end{minipage}
    \begin{minipage}{0.4\linewidth}
    \centering
        \begin{tabular}{lcc}
            \toprule
            Score &  Window Size &  F1 \\
            \midrule
            ip &            3 &    0.5119 \\
            ip &            5 &    0.5121 \\
            ip &            8 &    0.5123 \\
            \midrule
            cb &            3 &    0.5121 \\
            cb &            5 &    0.5122 \\
            cb &            8 &    0.5124 \\
            \midrule
            pt &            3 &    0.5059 \\
            pt &            5 &    0.5059 \\
            pt &            8 &    0.5062 \\
            \midrule
            sc &            3 &    0.5121 \\
            sc &            5 &    0.5122 \\
            sc &            8 &    0.5124 \\
            \midrule
            un &            3 &    0.5075 \\
            un &            5 &    0.5075 \\
            un &            8 &    0.5076 \\
            \midrule
            ctrs &            3 &    0.5077 \\
            ctrs &            5 &    0.5078 \\
            ctrs &            8 &    0.5080 \\
        \bottomrule
        \end{tabular}
        \end{minipage}
\caption{Bootstrapped baselines based by window size. Each value represents $3\sigma$ significance or 3 standard deviations above the mean.}
\label{tab:local_three_sigmas}
\end{table}

\begin{table}[ht!]
\small
    \centering
    \begin{tabular}{lcc}
\toprule
    Score   &  $2\sigma$ &  $3\sigma$    \\
\midrule
    ag      &     0.6268 &       0.6929 \\
    ap     &     0.6255 &       0.6914 \\
    co     &     0.6277 &       0.6935 \\
    fb      &     0.6209 &       0.6861 \\
    gd      &     0.6288 &       0.6950 \\
    hw      &     0.6256 &       0.6911 \\
    ip      &     0.6269 &       0.6925 \\
    cb      &     0.6315 &       0.6983 \\
    pt      &     0.6195 &       0.6848 \\
    sc     &     0.6273 &       0.6933 \\
    un      &     0.6246 &       0.6904 \\
    ctrs    &     0.6217 &       0.6871 \\  
\bottomrule
\end{tabular}

    \caption{Baselines for session level codes}
    \label{tab:global_3_sigmas}
\end{table}

\subsection{Windowed Dynamic Mode Decomposition}
\label{sec:windowed_dmd}
To represent conversational flow, we define the interaction between the therapist and client as a control-affine dynamical system. More concretely, taking the language of the client as an observation of a hidden state, influenced by an input signal consisting of the therapists language allows us represent the dynamics as a matrix couple consisting of the transitions and controls. 

Dynamic Mode Decomposition (DMD) is a method for approximating a linear dynamical system model from observable data \citep{schmid2010dynamic}. \citeauthor{proctor2016dynamic} built upon this work by extending the same underlying principle of fitting a DMD model would allow for a control input (DMDc), which can be trivially applied to the work done by \citet{zhang2019online} which demonstrated how a windowed and online implementation of DMD can be applied in order to more adequately account for non-stationary processes \citep{proctor2016dynamic, zhang2019online}. 

More explicitly, the extracted turn-level sentence embeddings outlined in \S \ref{sec:data} are constructed into arrays, $Y = [y_0, y_1, \hdots, y_T]$ and $X = [x_0, x_1, \hdots, x_T]$, for the client and therapist respectively over the session of length $T$. Our approach models the window of turns $Y_{w, t} = [y_t, y_{t+1}, \hdots y_{t+w}]$ and $X_{w,t} = [x_t, x_{t+1}, \hdots x_{t+w}]$ using windowed DMDc, represented by:
\begin{equation}
    Y_{t+1,w} = A_{t,w}Y_{t,w} + B_{t,w}X_{t,w} = \begin{bmatrix}A_{w, t} & B_{w, t}\end{bmatrix} \begin{bmatrix}
Y_{w, t} \\
X_{w, t}
    \end{bmatrix}
    \label{eq:windowed_dmdc}
\end{equation}

The dynamics over the window are captured by $A_{w, t}$ and $B_{w, t}$ in Eq \ref{eq:windowed_dmdc} which can be reconstructed by applying:
\begin{equation}
    \begin{bmatrix}A_{w, t} & B_{w, t}\end{bmatrix} = Y_{t+1,w}\begin{bmatrix}
Y_{w, t} \\
X_{w, t}
    \end{bmatrix}^{\dagger}
    \label{eq:windowed_dmdc_solution}
\end{equation}

By applying the Moore-Penrose pseudo-inverse, denoted by $\dagger$, the transition matrices $A_{w, t}$ and controller $B_{w, t}$ represent how the previous observations and the input signal approximately model the next time step.

To directly study the behavior of the dynamical model as outlined by these matrices, the eigenvalues, $\lambda \in \mathbb{C}$ can be extracted, which in turn can be interpreted as the $\mathcal{Z}$-transformed open-loop poles, which are the modes of the system \citep{ragazzini1952analysis}\footnote{The $\mathcal{Z}$-transform is a discrete-time analogue to the Laplace transform ($\mathcal{L}$) and the eigenvalues exist in the spectral space $z \in \mathbb{C}$ and correspond to the time-domain exponential that govern the system}. Through this lens, these eigenvalues can be interpreted as the complex exponential powers that describe the differential equations which describe the behavior observed in the temporal domain.

Briefly, the larger the magnitude of the eigenvalue the more ``dominant" the related mode is, implying that the related dynamics will persist for longer since the associated time-domain exponential is larger.\footnote{Typically we are interested in evaluating stable systems where the eigenvalues are contained by the unit circle and subsequently correspond to a dampened oscillator, however in the local dynamics extracted there were isolated moments when the system experiences instability. However in those cases the faster rising modes will still be dominant so the relationship between the magnitude and the dominance remains.} Extending this notion modes with associated $|\lambda| = 0$ dampen immediately and do not continue influence the evolution of the model. Using the imaginary component to compute the angle of deflection from the real axis, the angle $\omega$ is related to the frequency of the oscillatory response described in the mode. By including the eigenvalues of the transition matrix $(\lambda_T)$, it is possible to model the local influence the client's previous statements on their future statements, while the controller matrix eigenvalues ($\lambda_C$) contain information pertaining to the influence statements made by the therapist on the future statements of the client. In the experiments we study the information contained by $\lambda_T$ and $\lambda_C$ in isolation, as well as, combined $\lambda_{T+C}$.

\begin{figure}[ht!]
    \centering
    \includegraphics[width=0.99\linewidth]{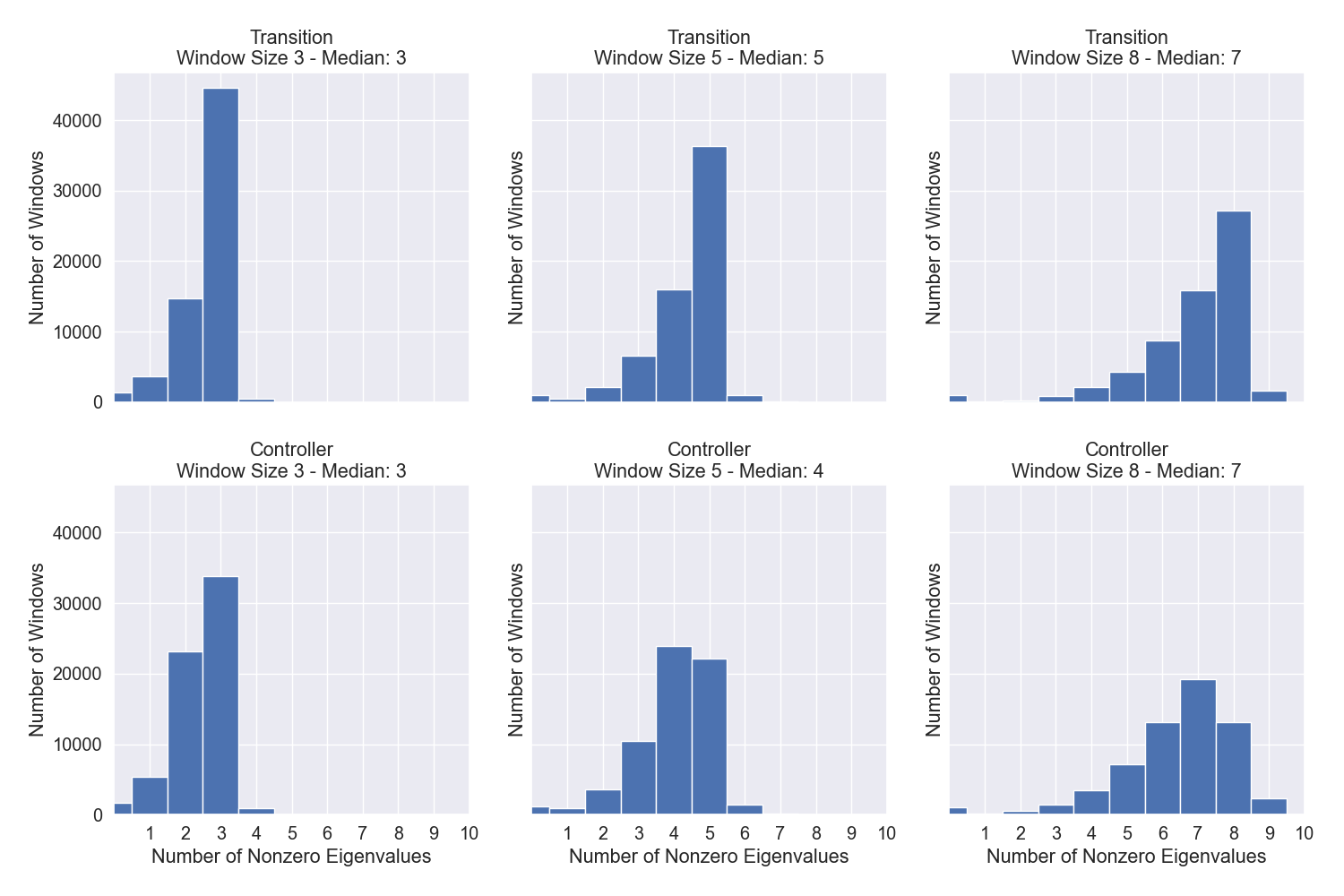}
    \caption{Number of non-zero eigenvalue for dynamics computed at varying window size for both transition (T) and controller (C)}
    \label{fig:nonzero-eig-distribution}
\end{figure}

Furthermore our experiments compare the use of different window sizes and include varying amounts of dominant eigenvalues extracted from the matrices. Specifically windows sizes, $w \in [3, 5, 8]$, were evaluated.\footnote{3 was chosen as 2 or fewer time-steps is too short a window to meaningfully capture the dynamics and 8 was a maximum to account for a session that was 9 turns long.} Similar evaluations were conducted to evaluate for include a number of eigenvalues. Experiments reflected in Figure \ref{fig:nonzero-eig-distribution}, enabled us to choose varying number of eigenvalues denoted by $n_{\lambda} \in [1, 3, 5, 7]$.

\begin{figure}[ht!]
    \centering
    \includegraphics[width=0.95\linewidth]{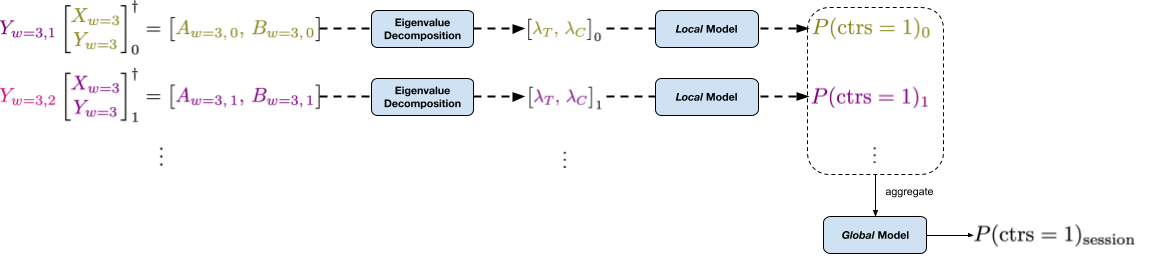}
    \caption{With the windows extracted using the pipeline outlined in Figure \ref{fig:window_extraction_pipeline} the windowed DMDc is applied in order to find the corresponding transition and controller matrices $[A, B]_t$ and their respective $[\lambda_T, \lambda_C]$. These are used to make \textit{local} predictions which are then accumulated and passed into the aggregate model for a \textit{global} prediction.}
    \label{fig:windows_to_ml_predictions_pipeline}
\end{figure}

\subsection{Models}
With the eigenvalues for each window size computed, a set of machine-learning models with varying parameters and underlying assumptions were trained to classify from each window for the \textit{local} prediction task.

The models trained included:
\begin{itemize}
    \item Gaussian Naive Bayes (GNB)
    \item Logistic Regression (LR)
    \item Linear Support Vector Machine $c \in [0.01, 0.1, 1, 10, 100]$ ($\text{L-SVM}_c$)
    \item $k$-nearest Neighbors $k = [1, 3, 5, 10, 30, 50, 100]$ ($\text{KNN}_k$)
\end{itemize}

Since our objective is to study whether these modes capture meaningful information about the CBT competence the models are kept relatively simple and a wide array are chosen to explore how different underlying assumptions about the data can lead to better local and global performances.

\subsubsection{Aggregating Global Scores from Local Predictions}
The models listed above are trained to make a prediction $P(score = 1 | \lambda)_{t}$, estimating the likelihood that the local dynamic modes belong to a CBT session that is high-competence. Since the data originates from a CBT competence score that is non-binary, it stands to reason that low-competence sessions (CTRS $< 40$) will have moments that will correspond to competence but not frequently enough to reach the threshold. For this reason if the \textit{local} predictions are more generally informative, accumulating them should yield a signal that is useful to predict the overall competence of the session, $P(score = 1)_\text{session}$.

Our analysis tests 2 different accumulation methods: \textit{sum} and \textit{average}. \textit{Sum} simply adds up all of the instances of competence over each of the extracted windows for a given session, while \textit{average} divides this value by the number windows that have been extracted from the session. Similarly, two different models for classification are used: \textit{training mean} (TM) and \textit{logistic regression} (LR). TM aggregates the mean over the training split and assigns a class 1 if the score for a test session is greater than or equal to the mean. LR uses the training split score to fit a simple linear model that predicts over the accumulated score for the test sessions.

\section{Results}
\label{sec:results}
All results below are averaged across a 5-fold cross validation. The splits are maintained for both of the \textit{local} and \textit{global} model training and accumulation. This splitting step helps account for erroneously good results that occur due to spuriously lucky splits of the data. The data splits are conditioned on the labels maintaining a relatively identical distribution of high- and low-competence sessions, as well as, accounting for the client, so that all sessions from the same client are kept together in a split, allowing us to know that the model is not over-fitting to anything about a specific client and their dialogue dynamics.

\subsection{Local Predictions}
Using the reference scores in Table \ref{tab:local_three_sigmas}, Table \ref{tab:local_results} displays the models that succeeded at having a significant ($3\sigma$) improvement over random chance. These models were the selected candidate models that were later applied for the global task. Notably, the best performing the models for each of the tasks use both eigenvalues from both the transition and controller (T+C) which suggests that it there is latent information in the dynamics of both the client and the affect of the therapist that meaningfully maps to CBT competence. 

GNB models seemed to have generally done the best, with $w = 8$ and $n_{\lambda} = 7$ which are the largest values possible. There are two components that likely contribute to this: 1. that the more windows there are the less sample there are for the model make mistakes on, 2. the longer range dynamics capture more meaningful representations with regards to the task of predicting CBT competence than those in smaller windows. This suggests that the time-varying component of the dynamics meaningfully changes over 8 talk turns or more.

Looking at the 3 scores which were the most effective locally: agenda (ag, 0.5749), key cognitive behavior (cb, 0.5608), and strategy for change (sc, 0.5624) stand out, suggesting that there is a more strongly observed signal in the extracted dynamics pertinent to those scores than to others . An interpretation that could account for this is that effective presentation of an agenda and strategy for change is referred to occur consistently throughout the session, similar to the dynamics of invoking key cognitive behavior.

The results in Table \ref{tab:local_results} suggest that these extracted eigenvalues contain a meaningful signal for the determination of CTRS scores and the competence of the session globally. Fundamentally, these models are not expected to perform particularly well as they are trying to predict a global label from only local information.  
Furthermore, the models do not ever observe the data directly, only a representation of the interaction of these signals over a small window sampled from the session. It reasons that high-competence moments exist in sessions that are scored overall low-competence. Further still, it is likely that the majority of moments are entirely neutral moments that the classifier would predict ambiguously (close to 0.5). 
% Our results suggest that there is something detectable in the way that clients and therapist engage that is specific to the competence of a particular therapist.
\begin{table}[ht!]
\centering
\begin{minipage}{0.45\linewidth}
\begin{tabular}{clcccc}
\toprule
                    Score &     Model & Input Type & $w$ &  $n_\lambda$ &  F1 \\
\midrule
                   ag &     $\text{L-SVM}_1$ &    T+C &            3 &         3 &    0.5267 \\
                   ag &   $\text{L-SVM}_{100}$ &      C &            5 &         5 &    0.5395 \\
                   ag &       GNB &    T+C &            8 &         7 &    0.5749 \\
\midrule
 ap &    $\text{L-SVM}_{10}$ &      C &            3 &         5 &    0.5163 \\
 ap &    $\text{L-SVM}_{10}$ &      C &            5 &         7 &    0.5301 \\
 ap &       GNB &    T+C &            8 &         7 &    0.5467 \\
 \midrule
            co &    $\text{KNN}_{50}$ &      T &            3 &         3 &    0.5169 \\
            co &   $\text{KNN}_{100}$ &      T &            5 &         3 &    0.5310 \\
            co &       GNB &    T+C &            8 &         7 &    0.5547 \\
\midrule
                 fb &    $\text{L-SVM}_{10}$ &      C &            3 &         7 &    0.5154 \\
                 fb &    $\text{L-SVM}_{10}$ &      C &            5 &         7 &    0.5222 \\
                 fb &       GNB &    T+C &            8 &         7 &    0.5401 \\ 
\midrule
         gd &    $\text{L-SVM}_{10}$ &      C &            3 &         7 &    0.5153 \\
         gd &    $\text{L-SVM}_{10}$ &      C &            5 &         5 &    0.5292 \\
         gd &       GNB &    T+C &            8 &         7 &    0.5355 \\
\midrule
                 hw &    $\text{L-SVM}_{10}$ &      C &            5 &         5 &    0.5134 \\
                 hw &       GNB &    T+C &            8 &         7 &    0.5303 \\
                 \hfill \\
            \bottomrule
\end{tabular}
\end{minipage}
\hspace{0.05\linewidth}
\begin{minipage}{0.45\linewidth}
\begin{tabular}{clcccc}
\toprule
                    Score &     Model & Input Type & $w$ &  $n_\lambda$ &  F1 \\

\midrule
            ip &    $\text{L-SVM}_{10}$ &      C &            3 &         1 &    0.5173 \\
            ip &     $\text{KNN}_{1}$ &      T &            5 &         5 &    0.5179 \\
            ip &     $\text{KNN}_{1}$ &    T+C &            8 &         7 &    0.5187 \\
\midrule
     cb &    $\text{L-SVM}_{10}$ &    T+C &            3 &         7 &    0.5249 \\
     cb &       GNB &    T+C &            5 &         5 &    0.5400 \\
     cb &       GNB &    T+C &            8 &         7 &    0.5608 \\
\midrule
          pt &    $\text{L-SVM}_{10}$ &      C &            3 &         3 &    0.5210 \\
          pt &    $\text{L-SVM}_{10}$ &      C &            5 &         5 &    0.5295 \\
          pt &       GNB &    T+C &            8 &         7 &    0.5514 \\
\midrule
      sc &     $\text{L-SVM}_1$ &      C &            3 &         3 &    0.5234 \\
      sc &   $\text{KNN}_{100}$ &      T &            5 &         5 &    0.5322 \\
      sc &       GNB &    T+C &            8 &         7 &    0.5624 \\
\midrule
            un &  $\text{L-SVM}_{0.01}$ &      C &            3 &         3 &    0.5075 \\
            un &    $\text{L-SVM}_{10}$ &    T+C &            5 &         5 &    0.5203 \\
            un &    $\text{L-SVM}_{10}$ &    T+C &            8 &         7 &    0.5338 \\
\midrule
                ctrs &    $\text{L-SVM}_{10}$ &      C &            3 &         5 &    0.5203 \\
                ctrs &    $\text{L-SVM}_{10}$ &    T+C &            5 &         5 &    0.5315 \\
                ctrs &       GNB &    T+C &            8 &         7 &    0.5514 \\
\bottomrule
\end{tabular}
\end{minipage}
\caption{Best performing models on local samples of segments by window size. Models with identical performance per a given window size}
\label{tab:local_results}
\end{table}

\subsection{Session-Level Predictions}
Table \ref{tab:global_results} lists the best performing models for each of the scores. When compared to the boot strap scores in Table \ref{tab:global_3_sigmas}, it's clear that most of the models outperform the 2$\sigma$ baselines but ultimately fall short of the 3$\sigma$. The homework (hw, 0.6191) and interpersonal (ip, 0.5995) do not globally perform better than the $2\sigma$ baseline.

The result of the interpersonal models does not come as much of a surprise since those local models were among the worst performing in Table \ref{tab:local_results} as well. Observing in Tables \ref{tab:local_three_sigmas} and \ref{tab:global_3_sigmas} that both the local and global baselines are each lower than many of the other scores suggests that this is generally a more difficult task, and that this method of accumulating and aggregating the scores from local dynamic windows is not adequate. Interestingly a separate Pearson correlation study showed that the model's interpersonal predictions correlated rather well ($r = 0.7595$, $p = 1.903 \times 10^{-39}$) on the training set but generalized extremely poorly to the test set ($r = 0.1303$, $p=0.4024$). By comparison the test set correlation for ctrs was $r = 0.4647$ with an associated $p = 0.0008$.\footnote{Traditionally correlations with $|r| > 0.3$ and $p < 0.05$ are considered significant} 

Notably, pacing and timing (pt, 0.6868) is the only score where the $3\sigma$ baseline is effectively met, with agenda (ag, 0.6835) and total CTRS (ctrs, 0.6767) coming close. According to the CTRS coding manual, the pacing and timing of a session evaluates the use of time via the induced structure and control the therapist exercises over the session. It is likely the case that setting an agenda at the beggining helps with the pacing and timing, which in turn strongly correlates with the overall CTRS competence. The result is particularly interesting result since this implies that the local dynamics extracted using the Windowed DMDc method and BERT embeddings capture have a meaningful interpretation of the interaction between the therapist and the client.

\begin{table}
\centering
\begin{tabular}{cccccccc}
\toprule
Score &   Model &  $n_{\lambda}$ & Input Type &  $w$ & Accumulator & Aggregator &     F1 \\
\midrule
ag &     GNB &              7 &        T+C &            8 &         sum &         LR & 0.6835 \\
ap &     GNB &              7 &        T+C &            8 &         sum &         LR & 0.6574 \\
co &     GNB &              7 &        T+C &            8 &         avg &         LR & 0.6470 \\
fb &     GNB &              7 &        T+C &            8 &         sum &         LR & 0.6580 \\
gd &     GNB &              7 &        T+C &            8 &         sum &         LR & 0.6577 \\
hw &     GNB &              7 &        T+C &            8 &         sum &         LR & 0.6191 \\
ip &   $\text{KNN}_1$ &              7 &        T+C &            8 &         avg &        TM & 0.5995 \\
cb &     GNB &              5 &        T+C &            5 &         avg &         LR & 0.6588 \\
pt &     GNB &              7 &        T+C &            8 &         avg &         LR & 0.6868 \\
sc &     GNB &              7 &        T+C &            8 &         avg &         LR & 0.6361 \\
un &  $\text{L-SVM}_{10}$ &              5 &        T+C &            5 &         avg &         LR & 0.6635 \\
ctrs &     GNB &              7 &        T+C &            8 &         sum &         LR & 0.6767 \\
\bottomrule
\end{tabular}
\caption{Models performing the best on the \textit{global} task}
\label{tab:global_results}
\end{table}

 Since the T+C models were found to be the most effective in predicting the session level competence it indicates that it is both the client's own internal transition, as well as, the way in which they interpret of the therapist's words that are most informative towards the predicting global scores. Most of these models were also using the largest window size ($w = 8$) and eigenvalues ($n_{\lambda}$ = 7) with the GNB model which is likely also a consequence of the results in Table \ref{tab:local_results} which down sampled the models that were used in the global predictions. 

Results were split fairly evenly between using sum and averages (avg) as the accumulation method, but the LR aggregator was clearly a stronger at separating the differences than using the simple training mean (TM).

\section{Conclusions}

\begin{figure}[ht]
    \centering
    \includegraphics[width=0.95\linewidth]{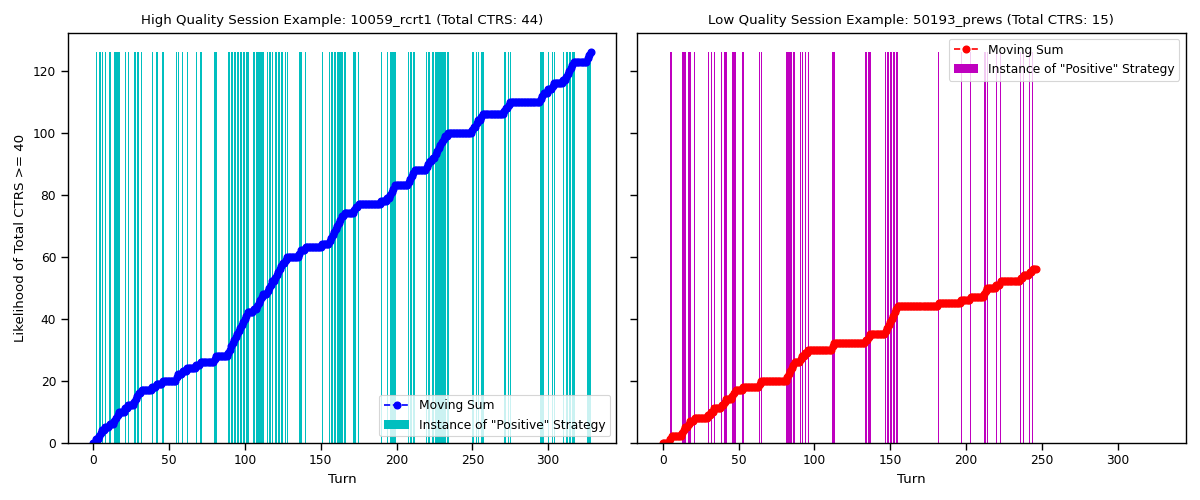}
    \caption{A sample trajectory of CTRS scores accumulating over the course for 2 sessions one being a high quality and one low quality session}
    \label{fig:sample_trajectory}
\end{figure}

Overall the results trend positively, suggesting there is significant value in extracting these dynamic modes and in their ability to adequately capture the desired behaviors of the therapist in assessing and providing feedback. 

This is particularly interesting when considering how these systems might be used in tracking and real-time feedback to the therapist. As seen in Figure \ref{fig:sample_trajectory} the resulting models can used to track the relative trajectory of the session as it is happening. With these models it would be possible to evaluate and even plan the future control signals. The fact that these systems perform well using a linear dynamical system assumption suggests that the computation and optimization steps would be relatively computationally inexpensive compared to alternative non-linear models.

\section{Future Work}
Our study demonstrates the value of the extracted dynamics, however the methods outlined for down stream classification are relatively simple and could likely be improved by the utilization of more complex sequence classification systems to make more complex predictions over time. Also training using a non-binary target might increase the separation of dynamic modes. Furthermore, it might be possible to improve overall prediction results by learning in a multi-task setting, as the sub-scores together may contain overlapping information that would improve individual sub-score predictions.

\citet{durante2022causal} demonstrated that measuring the error produced by the autonomous model and controlled model can be used to interpret the coordination between the individuals interacting which was found to be a useful towards predicting global measures about the interaction. In the future, it would possible to conduct a similar evaluation for this data and interpret those results in a similar way to the work done by \citet{martinez2019identifying} which looked at the alignment of narrative structures between clients and therapist to estimate alliance between the dyad, which is a related measure to a few of the scores of interest in CTRS.

Expanding the capabilities of these methods and building increasingly efficient automatic feedback mechanisms would allow for the development of concrete tools that therapists would be able to leverage to aid their clients. This is an exceptionally timely issue as the environments and platforms for delivering mental health care are especially challenged following the impacts of the recent global pandemic \citep{youn2020hidden, aknin2021mental}, as these challenges most heavily fall upon those systems most vulnerable and the least supported communities.

\bibliographystyle{abbrvnat}
\bibliography{references}

\end{document}